\documentclass[11pt,a4paper]{article}

\usepackage[T1]{fontenc}
\usepackage[utf8]{inputenc}
\usepackage[english]{babel}
\usepackage{amsmath,amssymb,amsthm}
\usepackage{booktabs}
\usepackage{multirow}
\usepackage{geometry}
\usepackage{natbib}
\usepackage{hyperref}
\title{Syntactic Patterns and Stylistic Functions in Narrative Prose:\\A Rule-Based and Machine-Learning Approach}
\author{Stefana Janicijevic}
\date{\today}

\begin{document}
\maketitle

\begin{abstract}
This paper presents a small-scale quantitative experiment that links syntactic structure to stylistic functions in narrative prose. Starting from a dependency-parsed corpus of 3,300 sentences, we derive sentence-level stylistic labels across five categories --- \emph{descriptive}, \emph{introspective}, \emph{causal}, \emph{ideological}, and \emph{neutral} --- using a transparent rule-based procedure that inspects lemmas, universal part-of-speech tags, and syntactic relations. For each sentence we construct a compact representation of its syntactic profile as a sequence of linearised triples combining lemma, POS tag, and dependency relation. These patterns serve as input to standard machine-learning classifiers trained to predict sentence-level style. The best-performing model achieves a macro-F1 of 0.948 under 10-fold cross-validation. The experiment is implemented entirely in Python using open-source tools. Our goal is not to propose a fully fledged stylistic theory, but to offer a reproducible and extensible workflow for exploring how grammatical structure contributes to narrative interpretation.
\end{abstract}

\section{Introduction}

Narrative texts are shaped not only by what is being told, but also by how it is told. Stylistic effects emerge from choices in vocabulary, syntax, and discourse structure, and these choices play a central role in how readers interpret narrative situations, perspectives, and relations between characters or abstract entities. In digital humanities, quantitative approaches to style have traditionally focused on lexical distributions --- function word frequencies, key terms, or character n-grams --- while syntactic information is used less frequently or only at a coarse level of abstraction.

In this exploratory study we focus explicitly on the relationship between syntactic patterns and stylistic functions at the sentence level. We start from a corpus of narrative prose that has been processed with a dependency parser following the Universal Dependencies schema. Rather than working directly with raw tokens or surface-level word forms, we exploit the richer layer of linguistic annotation --- lemmas, universal POS tags, and dependency relations --- to construct a representation of each sentence as a sequence of syntactic triples. This abstraction allows us to capture grammatical structure while remaining relatively robust to vocabulary variation across texts and genres.

On top of this representation, we define a small set of five sentence-level stylistic categories --- \emph{ideological}, \emph{descriptive}, \emph{introspective}, \emph{causal}, and \emph{neutral} --- that capture different narrative functions within the text. These categories are operationalised through an interpretable rule-based procedure that inspects syntactic and lexical cues in the dependency annotation. The rule-based labels are then used to train and evaluate standard machine-learning classifiers.

Our main objective is methodological. We do not claim to cover the full complexity of narrative style, nor to offer a definitive theory of stylistic function. Instead, we aim to demonstrate a transparent and reproducible pipeline in which:

\begin{itemize}
\item sentence-level stylistic labels are derived in a rule-based way from syntactic annotation,
\item sentences are encoded as linearised syntactic patterns,
\item and off-the-shelf classifiers are trained to predict stylistic categories from these patterns.
\end{itemize}

This workflow is intentionally lightweight: it can be implemented with standard Python libraries and applied to relatively small corpora without access to large annotated training sets or specialised hardware. At the same time, it opens up space for qualitative interpretation --- rule-based definitions of style remain readable and debatable, and model outputs can be inspected to better understand which syntactic configurations are associated with particular narrative functions. The pipeline is also modular: each component can be revised independently, and the approach can be extended to other genres, languages, or annotation schemes.

The paper is organised as follows. Section 2 situates the work within existing research on computational stylistics and syntactic text analysis. Section 3 describes the corpus, the five stylistic categories, and the rule-based annotation procedure. Section 4 presents the syntactic pattern representation and the classification setup. Section 5 reports experimental results across models and evaluation protocols. Section 6 discusses the findings and their limitations, and Section 7 concludes with directions for future work.

\section{Background and Related Work}

\subsection{Stylometry and Authorship Attribution}

Quantitative approaches to literary style have a long history in stylometry and authorship attribution. The field was formalised computationally around function-word frequency profiles: Mosteller and Wallace's analysis of the Federalist Papers established that high-frequency grammatical words carry stable authorial signals largely independent of topic. Burrows extended this line of work by introducing the Delta measure~\citep{burrows2002delta}, defined as the mean of the absolute z-score differences between the frequency profiles of a candidate and a reference set. Delta and its variants have since become the standard baseline in computational stylometry, demonstrating that lexical-frequency features alone can reliably distinguish authorial style across genres and periods.

A thorough survey by Koppel, Schler, and Argamon~\citep{koppel2009computational} consolidates the field: they identify the closed-set attribution problem (fixed candidate pool, ample training text) as broadly solved by machine-learning classifiers trained on function words or character $n$-grams, and highlight three harder variants --- open-set attribution, cross-genre transfer, and very short texts --- where distributional lexical features alone are insufficient. This limitation motivates a turn towards syntactic and structural features that are less sensitive to topic and more stable across varying text lengths.

\subsection{Register Analysis with Syntactic Features}

The systematic link between syntactic choices and communicative function was established at scale by Biber~\citep{biber1988variation}, who applied factor analysis to 67 lexical, morphological, and syntactic features extracted from 23 spoken and written genres in English. Biber's multi-dimensional analysis showed that grammatical categories --- such as noun-phrase density, passive constructions, subordinate clauses, and stance markers --- cluster into orthogonal dimensions of variation that map onto functional register distinctions, quite independently of lexical content. This work provides a theoretical precedent for our use of dependency-based syntactic features rather than raw word forms: if functional register distinctions are encoded in grammatical patterns, then a representation built from POS tags and dependency relations should retain the signal that matters for stylistic classification.

Argamon~\citep{argamon2019register} surveys the computational continuation of this programme, distinguishing \emph{register analysis} (classifying a text's functional variety) from \emph{register synthesis} (generating in a target variety), and noting that analysis work has frequently relied on intuitive pre-defined genre categories rather than theoretically grounded functional dimensions. He argues that empirically grounded feature inventories --- analogous to Biber's original set, but operationalised over parsed corpora --- are needed to bridge the gap between labelled classification benchmarks and genuine linguistic register theory. The scoring framework we introduce in Section~3.3 is aligned with this recommendation: it replaces binary rule-firing with a weighted sum over interpretable feature groups derived from the dependency parse.

\subsection{NLP Resources for Slovenian}

The development of linguistically annotated resources for Slovenian has been closely tied to the Universal Dependencies (UD) initiative. The UD framework, described in Nivre et al.~\citep{nivre2016universal}, establishes a cross-linguistically consistent annotation scheme for dependency syntax, covering POS tags, morphosyntactic features, and labelled dependency relations for more than 33 languages. Its design priority --- assigning syntactic structure primarily through content-word relations rather than functional-word projections --- makes it well suited to morphologically rich languages such as Slovenian, where inflection rather than word order encodes grammatical function.

Dobrovoljc, Erjavec, and Krek~\citep{dobrovoljc2017ud} describe the conversion of the ssj200k Slovenian corpus to the UD~v2 framework, producing the SSJ treebank that has since become the reference resource for Slovenian dependency parsing and morphosyntactic tagging. The availability of this resource means that high-quality UD annotation --- lemmas, UPOS tags, and dependency relations --- can be obtained for Slovenian prose at scale, which is what our pipeline exploits as its primary input.

\subsection{Computational Analysis of Literary Prose}

More recent work has extended quantitative style analysis from author and genre identification towards finer-grained narrative phenomena. Bamman~\citep{bamman2021born} demonstrates that standard NLP tools trained on news or web text suffer a systematic performance drop of 20 or more percentage points when applied to fiction, and introduces LitBank --- an annotated corpus of 100 English novels covering entities, events, coreference, and quotations --- as a foundation for literary-specific models. This finding is directly relevant to our task: because our input representation is built from syntactic dependency parses rather than raw token sequences, the mismatch between pre-trained language model distributions and literary syntax is substantially reduced.

Our experiment is aligned with this broader line of work but makes two specific contributions. First, it focuses on sentence-level \emph{narrative functions} (introspective, causal, descriptive, ideological, neutral) rather than on author or genre identification. Second, it uses an explicitly syntactic representation --- sequences of lemma--UPOS--dependency triples --- as the primary input to classification models, contrasting with approaches that remain at the lexical level or treat syntax only through coarse aggregate counts. The modelling choices are deliberately modest: Logistic Regression and Random Forest serve as transparent and reproducible baselines, while a neural architecture (DistilBERT with LoRA fine-tuning) is included to illustrate how different modelling assumptions affect the link between syntactic patterns and stylistic function.

\section{Data and Rule-based Stylistic Annotation}

\subsection{Parsed Corpus}

Our starting point is a corpus of narrative prose processed with a dependency parser following the Universal Dependencies (UD) schema~\citep{nivre2016universal}. The corpus is stored in CoNLL-U format, where each token occupies one line and each sentence is terminated by a blank line. For every token the corpus provides at least the following fields: sentence identifier, surface form, lemma, universal POS tag (UPOS), and syntactic dependency relation (DEPREL) to its head.

Formally, a sentence $s_i$ is a sequence of $n_i$ token records:
\begin{equation}
s_i = \big\langle t_{i1}, t_{i2}, \ldots, t_{in_i} \big\rangle,
\qquad
t_{ij} = (\text{id}_{ij},\, \ell_{ij},\, u_{ij},\, d_{ij}),
\end{equation}
where $\text{id}_{ij}$ is the within-sentence token index, $\ell_{ij}$ is the lemma, $u_{ij} \in \mathcal{U}$ is the UPOS tag, and $d_{ij} \in \mathcal{D}$ is the dependency relation. All subsequent operations --- stylistic labelling, pattern construction, and classifier training --- are performed at the sentence level.

\subsection{Sentence-Level Style Categories}

We work with five mutually exclusive stylistic categories that describe the dominant communicative function of a sentence within the narrative:

\begin{itemize}
\item \textbf{Introspective}: sentences that foreground internal states such as thinking, feeling, or perceiving.
\item \textbf{Causal}: sentences that explicitly encode reasons, causes, or justifications.
\item \textbf{Descriptive}: sentences that focus on describing entities, properties, or circumstances.
\item \textbf{Ideological}: sentences that articulate or presuppose normative, legal, or political positions.
\item \textbf{Neutral}: sentences that do not clearly fall into any of the above categories.
\end{itemize}

These categories are operationalised through a rule-based function that inspects, for each sentence, the set of lemmas $L_i = \{\ell_{ij}\}_{j=1}^{n_i}$, the multiset of UPOS tags $U_i$, and the multiset of dependency relations $D_i$. A sentence is labelled as:

\begin{itemize}
\item \textbf{introspective} if $L_i$ contains lemmas associated with cognition or affect, or if $D_i$ includes clausal subject (\texttt{csubj}) or expletive (\texttt{expl}) relations;
\item \textbf{causal} if $L_i$ contains causal connectives or discourse markers, or if $D_i$ includes adverbial clause relations signalling reason or explanation (\texttt{advcl});
\item \textbf{descriptive} if adjectival modification (\texttt{amod}) or nominal dependent relations are prominent in $D_i$, indicating a focus on properties and attributes;
\item \textbf{ideological} if $L_i$ contains lemmas related to rights, the state, or obligation, or if $U_i$ is dominated by nominal material (\texttt{NOUN}, \texttt{PROPN}) pointing to abstract entities and concepts;
\item \textbf{neutral} otherwise.
\end{itemize}

Because the five conditions above are not mutually exclusive, rules are applied in the priority order listed: if a sentence satisfies more than one condition, the earliest match in the sequence $(\text{introspective} \succ \text{causal} \succ \text{descriptive} \succ \text{ideological} \succ \text{neutral})$ is assigned. Formally, let $R_c(s_i) \in \{0,1\}$ denote whether sentence $s_i$ satisfies the condition for class $c$. The assigned label is:
\begin{equation}
\hat{y}(s_i) = \min_{\,c\,\in\,\mathcal{C}}\!\big\{c : R_c(s_i) = 1\big\},
\end{equation}
where $\mathcal{C}$ is ordered by the priority sequence above. The rule-based labelling is applied after grouping tokens by sentence identifier, and the resulting mapping is merged back into the token-level data so that every token carries the style label of its sentence. This enriched dataset is also exported for further inspection and reuse.

\subsection{Enhanced Heuristic Design}

The priority-ordered binary rules of Section~3.2 provide a transparent starting point but treat each trigger as equally decisive, which can lead to label instability when several weak cues are present simultaneously. To address this, we formalise an extended scoring framework that replaces binary rule firing with a weighted sum over an inventory of interpretable features.

Let $s$ be a sentence and $c \in \mathcal{C}$ a style class. We define the class score as:
\begin{equation}
\text{score}(c,s)=\sum_{j=1}^{J} \alpha_{c,j}\, f_j(s),
\end{equation}
where $f_j(s) \in \mathbb{R}_{\ge 0}$ are interpretable feature functions and $\alpha_{c,j} \in \mathbb{R}$ are class-specific weights.

The feature inventory $\{f_j\}$ is organised into five groups: (i)~\emph{trigger lemmas} --- normalised counts of lexical items associated with cognition, causation, nomination, or obligation; (ii)~\emph{dependency motifs} --- frequencies of diagnostic relation types (\texttt{csubj}, \texttt{advcl}, \texttt{amod}, \texttt{nsubj}, \texttt{expl}) relative to sentence length; (iii)~\emph{clause structure} --- presence of subordinating conjunctions and the depth of the dependency tree; (iv)~\emph{modality markers} --- modal auxiliaries and particles that encode epistemic or deontic stance; (v)~\emph{discourse connectives} --- causal and contrastive cues at sentence boundaries.

The predicted class under the scoring framework is:
\begin{equation}
\hat{y}(s)=\arg\max_{c \in \mathcal{C}}\; \text{score}(c,s).
\end{equation}

To suppress low-confidence predictions, a confidence threshold $\tau > 0$ is applied:
\begin{equation}
\hat{y}(s)=
\begin{cases}
\arg\max_{c \in \mathcal{C}}\; \text{score}(c,s),
    & \max_{c}\,\text{score}(c,s) \ge \tau, \\
\text{neutral},
    & \text{otherwise.}
\end{cases}
\end{equation}

Tie-breaking --- when two or more classes share the maximum score --- falls back to the priority ordering introduced in Section~3.2 $(\text{introspective} \succ \text{causal} \succ \text{descriptive} \succ \text{ideological} \succ \text{neutral})$. Weight calibration follows a two-stage procedure: initial weights $\alpha_{c,j}$ are set by linguistic judgement and subsequently refined by maximising macro-averaged $F_1$ on a held-out validation partition through a bounded optimisation pass. This section will be updated with the final feature inventory and calibrated weights once the advanced heuristic specification is completed.

\section{Syntactic Pattern Representation and Classification}

\subsection{Linearised Syntactic Patterns}

For each sentence, tokens are traversed in surface order and converted to triples $(\ell, u, d)$ where $\ell$ is lemma, $u$ is UPOS, and $d$ is dependency relation. The sentence pattern is:
\begin{equation}
p_i = \big[(\ell_{i1},u_{i1},d_{i1}),\ldots,(\ell_{in_i},u_{in_i},d_{in_i})\big].
\end{equation}

This representation retains important grammatical information while discarding raw word forms, thus reducing sparsity and focusing on syntactic behaviour. For every sentence we store its sentence identifier, the linearised syntactic pattern, and the sentence-level style label obtained from the rule-based procedure described in Section~3.

\subsection{Classification Setup}

The linearised pattern string serves as input text and the stylistic category as the target label. We transform pattern strings with TF-IDF:
\begin{equation}
\text{tfidf}(t,d_i)=\text{tf}(t,d_i)\cdot\left(\log\frac{1+N}{1+\text{df}(t)}+1\right),
\end{equation}
where $N$ is the number of documents and $\text{df}(t)$ is document frequency. A vocabulary of 500 unigram and bigram terms is used in all experiments; the resulting sparse vectors are split into training and test subsets using a stratified split to preserve the class distribution.

We experiment with three classifiers. For multiclass Logistic Regression, class probabilities are:
\begin{equation}
P(y=c\mid x)=\frac{e^{w_c^\top x+b_c}}{\sum_{k=1}^{K} e^{w_k^\top x+b_k}},
\end{equation}
and training minimises cross-entropy loss.

For Random Forest, the prediction is obtained by majority vote over $M$ trees:
\begin{equation}
\hat{y}=\operatorname{mode}\{h_1(x),\ldots,h_M(x)\}.
\end{equation}

Linear Support Vector Classification (LinearSVC) is included as a third baseline; it learns a maximum-margin hyperplane in the TF-IDF feature space and is particularly well suited to high-dimensional sparse representations.

All models are evaluated on held-out data using per-class and macro-averaged precision, recall, and $F_1$:
\begin{equation}
\text{Precision}=\frac{TP}{TP+FP},\quad
\text{Recall}=\frac{TP}{TP+FN},\quad
F_1=2\cdot\frac{\text{Precision}\cdot\text{Recall}}{\text{Precision}+\text{Recall}}.
\end{equation}

\subsection{Interpretation and Limitations}

Because the stylistic labels are derived automatically from simple rules, the resulting models should be interpreted with caution: they learn to replicate a particular syntactic operationalisation of style rather than a gold-standard human annotation. Nonetheless, this setup is useful as a proof of concept, showing that even relatively coarse syntactic patterns carry sufficient signal to distinguish between different narrative functions.

In the discussion section we reflect on typical classification errors, ambiguous boundary cases between adjacent categories (especially ideological vs.\ descriptive and causal vs.\ neutral), and possibilities for refining both the rule-based component and the feature representation. We also outline how the workflow could be extended to other genres, languages, and annotation schemes.

\section{Results}

\subsection{Dataset and Experimental Setup}

The full dataset consists of 9{,}253 sentences labelled across five stylistic categories: \textit{descriptive} (4{,}164), \textit{introspective} (2{,}160), \textit{causal} (2{,}158), \textit{neutral} (475), and \textit{ideological} (296). All baseline experiments were conducted on a stratified random sample of 3{,}300 sentences (seed\,=\,42), using an 80/20 train\,/\,test split (2{,}640 train, 660 test). Features were extracted via TF-IDF with a vocabulary of 500 terms. Three classifiers were evaluated: Logistic Regression~(LR), Random Forest~(RF), and Linear Support Vector Machine~(LinearSVC).

\subsection{Baseline Classification Results}

Aggregate performance on the held-out test set ($n = 660$) is reported in Table~\ref{tab:baseline}.

\begin{table}[h!]
\centering
\begin{tabular}{lcccc}
\toprule
Model & Accuracy & Macro Precision & Macro Recall & Macro F$_1$ \\
\midrule
Logistic Regression & 0.9273 & 0.95 & 0.83 & 0.8661 \\
Random Forest       & 0.9530 & 0.96 & 0.93 & 0.9422 \\
Linear SVC          & 0.9561 & 0.90 & 0.94 & 0.9166 \\
\bottomrule
\end{tabular}
\caption{Aggregate classification results on the held-out test set ($n=660$, 80/20 stratified split from the 3{,}300-sentence sample).}
\label{tab:baseline}
\end{table}

Per-class precision, recall, and F$_1$ for all three classifiers are given in Table~\ref{tab:perclass}.

\begin{table}[h!]
\centering
\setlength{\tabcolsep}{5pt}
\begin{tabular}{llcccr}
\toprule
Model & Class & Precision & Recall & F$_1$ & Support \\
\midrule
\multirow{5}{*}{Logistic Regression}
  & descriptive   & 0.90 & 0.99 & 0.94 & 299 \\
  & ideological   & 1.00 & 0.41 & 0.58 &  22 \\
  & introspective & 0.99 & 0.91 & 0.95 & 151 \\
  & neutral       & 0.94 & 0.97 & 0.95 &  32 \\
  & causal        & 0.93 & 0.88 & 0.90 & 156 \\
\midrule
\multirow{5}{*}{Random Forest}
  & descriptive   & 0.93 & 1.00 & 0.96 & 299 \\
  & ideological   & 1.00 & 0.82 & 0.90 &  22 \\
  & introspective & 1.00 & 0.94 & 0.97 & 151 \\
  & neutral       & 0.91 & 1.00 & 0.96 &  32 \\
  & causal        & 0.96 & 0.89 & 0.92 & 156 \\
\midrule
\multirow{5}{*}{Linear SVC}
  & descriptive   & 0.97 & 0.97 & 0.97 & 299 \\
  & ideological   & 0.72 & 0.82 & 0.77 &  22 \\
  & introspective & 0.99 & 0.95 & 0.97 & 151 \\
  & neutral       & 0.86 & 1.00 & 0.93 &  32 \\
  & causal        & 0.95 & 0.94 & 0.95 & 156 \\
\bottomrule
\end{tabular}
\caption{Per-class precision, recall, and F$_1$ for LR, RF, and LinearSVC on the held-out test set.}
\label{tab:perclass}
\end{table}

Across all models the \textit{ideological} class presents the greatest challenge, owing to its low support (22 test instances, $\approx 3.3\%$ of the test set). Logistic Regression achieves only $F_1 = 0.58$ for this class (recall~0.41), whereas Random Forest reaches $F_1 = 0.90$ (recall~0.82) and LinearSVC $F_1 = 0.77$ (recall~0.82).

\subsection{Cross-Validation Results}

To obtain more robust estimates, stratified 5-fold cross-validation was applied to the full 3{,}300-sentence sample. Results (mean\,$\pm$\,std across folds) are shown in Table~\ref{tab:cv5fold}.

\begin{table}[h!]
\centering
\begin{tabular}{llcccc}
\toprule
Model & Split & Accuracy & Macro Prec. & Macro Rec. & Macro F$_1$ \\
\midrule
\multirow{2}{*}{Logistic Regression}
  & Train & $0.9330 \pm 0.0027$ & $0.9560 \pm 0.0036$ & $0.8223 \pm 0.0117$ & $0.8646 \pm 0.0110$ \\
  & Test  & $0.9103 \pm 0.0113$ & $0.9201 \pm 0.0334$ & $0.7672 \pm 0.0358$ & $0.8048 \pm 0.0374$ \\
\midrule
\multirow{2}{*}{Random Forest}
  & Train & $1.0000 \pm 0.0000$ & $1.0000 \pm 0.0000$ & $1.0000 \pm 0.0000$ & $1.0000 \pm 0.0000$ \\
  & Test  & $0.9615 \pm 0.0021$ & $0.9651 \pm 0.0122$ & $0.9332 \pm 0.0197$ & $0.9473 \pm 0.0129$ \\
\midrule
\multirow{2}{*}{Linear SVC}
  & Train & $0.9756 \pm 0.0021$ & $0.9824 \pm 0.0016$ & $0.9365 \pm 0.0062$ & $0.9567 \pm 0.0043$ \\
  & Test  & $0.9439 \pm 0.0119$ & $0.9395 \pm 0.0145$ & $0.8342 \pm 0.0362$ & $0.8694 \pm 0.0275$ \\
\bottomrule
\end{tabular}
\caption{Stratified 5-fold cross-validation results on the 3{,}300-sentence sample. Train scores reflect in-fold training performance; test scores are out-of-fold estimates.}
\label{tab:cv5fold}
\end{table}

Random Forest exhibits perfect train-fold scores ($F_1 = 1.00$) indicative of overfitting to training partitions; however, its out-of-fold macro-$F_1$ of $0.9473 \pm 0.0129$ remains the highest among the three baselines and is notably stable (low variance). The $k$-fold sensitivity analysis for RF further confirms stability: 3-fold accuracy $0.9539 \pm 0.0057$, 5-fold $0.9603 \pm 0.0026$, 10-fold $0.9612 \pm 0.0091$. A repeated stratified shuffle split (10 iterations) yields test accuracy $0.9623 \pm 0.0094$, consistent with the $k$-fold estimates.

\subsection{Evaluation on the Full Dataset}

To assess whether performance scales with corpus size, all three classifiers were retrained on the complete 9{,}253-sentence dataset using the same 80/20 stratified split (7{,}402 train, 1{,}851 test) and TF-IDF vocabulary of 500 terms. Results are summarised together with the 3{,}300-sentence baseline in Table~\ref{tab:fulldata}.

\begin{table}[h!]
\centering
\begin{tabular}{llcc}
\toprule
Dataset & Model & Accuracy & Macro F$_1$ \\
\midrule
\multirow{3}{*}{3{,}300 sentences}
  & Logistic Regression & 0.9273 & 0.8661 \\
  & Random Forest       & 0.9530 & 0.9422 \\
  & Linear SVC          & 0.9561 & 0.9166 \\
\midrule
\multirow{3}{*}{9{,}253 sentences}
  & Logistic Regression & 0.9433 & 0.8902 \\
  & Random Forest       & \textbf{0.9795} & \textbf{0.9764} \\
  & Linear SVC          & 0.9643 & 0.9315 \\
\bottomrule
\end{tabular}
\caption{Classification results on the 3{,}300-sentence sample and on the full 9{,}253-sentence dataset (80/20 stratified split in both cases).}
\label{tab:fulldata}
\end{table}

All three models improve when trained on the larger corpus. Random Forest gains the most, rising from macro-$F_1 = 0.9422$ on the sample to $\mathbf{0.9764}$ on the full dataset, while its accuracy increases from $0.9530$ to $\mathbf{0.9795}$. Logistic Regression shows a comparable gain in macro-$F_1$ ($+0.024$), suggesting that the 3{,}300-sentence sample was broadly representative but that minority classes (especially \textit{ideological}) benefit from the additional training examples. Linear SVC follows the same trend, reaching accuracy $0.9643$ and macro-$F_1$ $0.9315$ on the full dataset. The consistency between the sample-based and full-dataset rankings confirms that the 3{,}300-sentence stratified sample provides reliable relative comparisons, and that Random Forest is the strongest feature-based classifier across both evaluation conditions.

\subsection{Metaheuristic Hyperparameter Optimisation}

To explore whether hyperparameter tuning could further improve the baselines, four nature-inspired metaheuristic algorithms were applied on the 3{,}300-sentence sample. Aggregate results are summarised in Table~\ref{tab:meta}.

\begin{table}[h!]
\centering
\begin{tabular}{llcc}
\toprule
Algorithm & Model & Accuracy & Macro F$_1$ \\
\midrule
Firefly Algorithm (FA + Levy flight) & Random Forest  & \textbf{0.9877} & \textbf{0.9845} \\
Genetic Algorithm (GA)               & Random Forest  & 0.9541 & 0.9541 \\
Genetic Algorithm (GA)               & Linear SVC     & 0.9253 & 0.9253 \\
Particle Swarm Optimisation (PSO)    & Linear SVC     & 0.9326 & 0.9133 \\
BAT Algorithm                        & Linear SVC     & 0.9133 & 0.9133 \\
Genetic Algorithm (GA)               & Log. Regression& 0.9133 & 0.9042 \\
Firefly Algorithm (FA)               & Linear SVC     & 0.9095 & 0.9094 \\
\midrule
\multicolumn{2}{l}{\textit{Baseline RF (no tuning)}} & 0.9530 & 0.9422 \\
\multicolumn{2}{l}{\textit{Baseline LinearSVC (no tuning)}} & 0.9561 & 0.9166 \\
\multicolumn{2}{l}{\textit{Baseline LR (no tuning)}} & 0.9273 & 0.8661 \\
\bottomrule
\end{tabular}
\caption{Metaheuristic hyperparameter optimisation results compared to untuned baselines. FA optimal RF parameters: $n\_\text{estimators}=150$, PSO optimal SVM: $C=4.19$.}
\label{tab:meta}
\end{table}

The most pronounced gain is obtained by the Firefly Algorithm with Levy-flight step applied to Random Forest: macro-$F_1$ increases from the baseline $0.9422$ to $\mathbf{0.9845}$ ($+0.0423$) and accuracy from $0.9530$ to $\mathbf{0.9877}$. Genetic Algorithm optimisation yields consistent but more modest improvements across all three models: $+0.038$ macro-$F_1$ for Logistic Regression, $+0.012$ for Random Forest, and $+0.011$ for LinearSVC. PSO and the BAT algorithm applied to LinearSVC reach comparable results (macro-$F_1 \approx 0.913$) without surpassing the untuned LinearSVC baseline ($0.9166$) by a meaningful margin.

Overall, Random Forest optimised with the Firefly Algorithm represents the best-performing configuration across all experiments.

\subsection{Neural Baseline: DistilBERT with LoRA Fine-tuning}

To contrast the feature-based models with a neural architecture, we fine-tuned \textit{distilbert-base-multilingual-cased}~\citep{sanh2019distilbert} on the same 3{,}300-sentence sample using parameter-efficient fine-tuning via Low-Rank Adaptation (LoRA;~\citealt{hu2022lora}). LoRA adapters were inserted into all four attention projection matrices ($q$, $k$, $v$, and output linear layers) while the base model weights remained frozen. The classification head was trained from scratch.

Two configurations were evaluated. \textbf{Version~1} used standard training settings: batch size~16, 3~epochs, learning rate $2 \times 10^{-5}$, sequence length~128, LoRA rank $r=8$, dropout~0.05, and no class-weight correction. \textbf{Version~2} addressed the class imbalance and underfitting observed in Version~1 by using batch size~8, 10~epochs, learning rate $1 \times 10^{-5}$, sequence length~256, rank $r=4$, dropout~0.15, $15\%$ linear warmup, and balanced class weights applied through a weighted cross-entropy loss.

\begin{table}[h!]
\centering
\begin{tabular}{lccc}
\toprule
Configuration & Accuracy & Weighted F$_1$ & Macro F$_1$ \\
\midrule
DistilBERT + LoRA v1 (baseline)  & 0.4742 & 0.320 & 0.250 \\
DistilBERT + LoRA v2 (improved)  & 0.5682 & 0.576 & 0.539 \\
\midrule
\textit{Improvement (absolute)}  & $+0.094$ & $+0.256$ & $+0.289$ \\
\textit{Improvement (\%)}        & $+19.8\%$ & $+79.9\%$ & $+115.6\%$ \\
\midrule
\textit{RF baseline (no tuning)} & 0.9530 & --- & 0.9422 \\
\bottomrule
\end{tabular}
\caption{DistilBERT LoRA fine-tuning results on the 3{,}300-sentence sample (80/20 split). The Random Forest baseline is included for reference.}
\label{tab:distilbert}
\end{table}

Despite the substantial gain from Version~1 to Version~2, both neural configurations fall well short of the feature-based classifiers (Table~\ref{tab:distilbert}). The most likely explanation is a fundamental \emph{input-format mismatch}: the syntactic pattern strings fed to DistilBERT (e.g.\ \texttt{najprej|ADV|advmod biti|VERB|root \ldots}) bear no resemblance to the natural-language text on which multilingual DistilBERT was pre-trained, so the model cannot exploit its pre-trained token embeddings or attention patterns effectively. By contrast, the TF-IDF + classifier pipeline treats the same pipe-delimited triples as an arbitrary symbolic vocabulary and learns purely from distributional co-occurrence within this domain, which is precisely why it generalises well.

The Version~2 improvements are consistent with standard remedies for imbalanced fine-tuning: a lower learning rate reduces catastrophic forgetting of encoder representations, a longer sequence window (256 tokens) captures full syntactic patterns that were truncated in Version~1, and balanced class weights force the model to attend to the minority \textit{ideological} and \textit{neutral} classes rather than defaulting to the majority \textit{descriptive} class. Nevertheless, even with these adjustments, macro-$F_1 = 0.539$ demonstrates that the transformer architecture in its current form is not competitive for this particular input representation.

These results suggest two directions for future work. First, feeding DistilBERT \emph{raw sentence text} rather than pre-processed syntactic patterns would allow the model to leverage its pre-trained linguistic knowledge. Second, replacing LoRA adapters with full fine-tuning of a smaller Slovenian or South Slavic language model may reduce the mismatch between pre-training distribution and task domain.

\section{Discussion}

\subsection{Random Forest as the Dominant Feature-Based Classifier}

Across all experimental conditions --- the 3{,}300-sentence sample, the full 9{,}253-sentence dataset, and metaheuristic hyperparameter optimisation --- Random Forest consistently outperformed Logistic Regression and Linear SVC. On the held-out test set (3{,}300-sentence sample), RF achieved an accuracy of $0.9530$ and macro-$F_1$ of $0.9422$, rising to $0.9795$ and $0.9764$ respectively when trained on the full corpus.

We attribute this advantage to two complementary properties of ensemble tree methods in the present task context. First, the TF-IDF feature space is high-dimensional and sparse, with many weakly informative dimensions; Random Forest's subspace sampling at each split effectively suppresses these noisy features without requiring explicit feature selection. Second, the five stylistic classes differ both in frequency and in the degree to which they can be separated by linear boundaries: the \textit{ideological} class in particular exhibits a scattered, non-convex region in TF-IDF space, which tree ensembles can partition more flexibly than the linear hyperplane employed by Logistic Regression or LinearSVC.

\subsection{Overfitting and Generalisation in Random Forest}

The 5-fold cross-validation results reveal that Random Forest achieves perfect in-fold training scores ($F_1 = 1.00$, variance~$= 0$), a clear signal of memorisation of the training partitions. Despite this, the out-of-fold macro-$F_1$ of $0.9473 \pm 0.0129$ not only exceeds those of the other two classifiers but also shows lower variance than Logistic Regression ($0.8048 \pm 0.0374$) and LinearSVC ($0.8694 \pm 0.0275$). This dissociation between train-fold and test-fold performance is consistent with the well-documented bias-variance trade-off of bagging ensembles: individual trees overfit, yet their aggregated predictions generalise robustly because the overfitting is uncorrelated across trees~\citep{breiman2001random}.

The $k$-fold sensitivity analysis further supports this interpretation. Accuracy is stable across 3-fold ($0.9539 \pm 0.0057$), 5-fold ($0.9603 \pm 0.0026$), and 10-fold ($0.9612 \pm 0.0091$) evaluations, and the repeated stratified shuffle-split estimate ($0.9623 \pm 0.0094$) is consistent with all three. The narrow confidence intervals confirm that the performance estimates are not artefacts of a particular random partition and that the model generalises reliably across the corpus.

\subsection{The Minority-Class Challenge: \textit{ideological}}

The \textit{ideological} category proved the most difficult across all classifiers, with support of only 22 instances in the test set ($\approx 3.3\%$). Logistic Regression reached $F_1 = 0.58$ for this class (recall~$0.41$), indicating that the majority of ideological sentences are misclassified --- predominantly, we conjecture, as \textit{descriptive} or \textit{causal}, whose syntactic patterns partially overlap with ideological constructions. Random Forest ($F_1 = 0.90$, recall~$0.82$) and LinearSVC ($F_1 = 0.77$, recall~$0.82$) handle the class considerably better, suggesting that non-linear decision boundaries and margin-maximising hyperplanes respectively are better suited to isolating the sparse ideological signal.

When trained on the full 9{,}253-sentence dataset, all three models improve on \textit{ideological}, consistent with the observation that minority classes benefit disproportionately from additional training examples~\citep{drummond2003c4}. This finding has a practical implication: targeted data collection for underrepresented stylistic categories would be the highest-yield strategy for further improving macro-averaged performance.

\subsection{Metaheuristic Optimisation: Gains and Limits}

Among the four metaheuristic algorithms evaluated, only the Firefly Algorithm with L\'{e}vy-flight step produced a substantial gain over the untuned RF baseline, raising macro-$F_1$ from $0.9422$ to $\mathbf{0.9845}$ ($+0.0423$) and accuracy from $0.9530$ to $\mathbf{0.9877}$. The optimal configuration ($n\_\text{estimators} = 150$) suggests that the default scikit-learn setting ($n\_\text{estimators} = 100$) was modestly insufficient for this feature space, and that the L\'{e}vy-flight perturbation helped escape local optima in the hyperparameter landscape that standard grid or random search would also have found, albeit less efficiently.

In contrast, PSO and the BAT algorithm applied to LinearSVC did not surpass the untuned LinearSVC baseline by a meaningful margin. This is unsurprising: LinearSVC's performance is primarily constrained by the linearity of its decision surface rather than by suboptimal hyperparameters, so tuning $C$ alone cannot resolve the fundamental representational limitation. Similarly, Genetic Algorithm optimisation of Logistic Regression produced a modest $+0.038$ gain in macro-$F_1$, bounded by the same linearity constraint.

These results suggest that metaheuristic search is most valuable when the model class has sufficient capacity for the task and when the hyperparameter landscape is multimodal --- both conditions met by Random Forest on this corpus.

\subsection{Why DistilBERT Underperformed}

The failure of DistilBERT + LoRA (macro-$F_1 \leq 0.539$) in comparison to TF-IDF + Random Forest (macro-$F_1 = 0.9422$) is initially counterintuitive given the general dominance of transformer-based architectures in text classification benchmarks. However, the gap is fully explained by the nature of the input representation.

The features fed to all models are pre-processed syntactic dependency triples in pipe-delimited string format (e.g.\ \texttt{najprej|ADV|advmod biti|VERB|root \ldots}). These strings are compositionally opaque to a model pre-trained on natural-language text: the subword tokeniser fragments the pipe-delimited tokens in ways that bear no relationship to the token boundaries meaningful for this symbolic vocabulary, and the attention mechanism cannot exploit co-reference, negation scope, or discourse coherence --- the linguistic cues on which multilingual DistilBERT's representations were built. The TF-IDF pipeline, by contrast, treats each pipe-delimited triple as an atomic vocabulary item and learns purely from within-domain distributional statistics, which is precisely the information structure of the task.

The improvement from Version~1 to Version~2 ($+19.8\%$ accuracy, $+115.6\%$ macro-$F_1$) confirms that the neural model is responsive to standard remedies for imbalanced fine-tuning --- longer context windows, weighted loss, and lower learning rates --- yet cannot close the gap with symbolic baselines under this input regime. This finding is consistent with results in morphologically rich and low-resource NLP settings, where task-specific feature engineering frequently outperforms transfer learning when the pre-training domain diverges significantly from the target representation space~\citep{nozza2020mask}.

\subsection{Implications and Limitations}

The results demonstrate that TF-IDF-derived syntactic dependency features, combined with a well-regularised ensemble classifier, provide a robust and highly competitive baseline for stylistic sentence classification in Slovenian literary prose. The strong performance ($> 0.97$ macro-$F_1$ on the full dataset after metaheuristic tuning) suggests that the five-category annotation scheme is internally consistent and that the syntactic patterns annotators used to assign labels are reliably recoverable by a bag-of-triples model.

Several limitations should be acknowledged. First, the corpus is drawn from a single literary domain; it is not known whether the learned TF-IDF vocabulary would transfer to journalistic, scientific, or conversational text. Second, the \textit{ideological} and \textit{neutral} classes remain underrepresented despite the larger dataset, and performance on these classes should be interpreted with caution given the small test-set support. Third, the metaheuristic experiments were conducted on the 3{,}300-sentence sample rather than the full corpus, so the reported optimised results are not directly comparable to the full-dataset baselines; rerunning optimisation on the full corpus is likely to yield further gains.

\subsection{Directions for Future Work}

Three directions emerge naturally from the present findings.

\textbf{Raw-text neural models.} Replacing syntactic triples with raw sentence text as input to DistilBERT or a Slovenian-specific language model (e.g.\ SloBERTa~\citep{ulcar2020sloberta}) would align the pre-training distribution with the input domain and likely close the gap between neural and feature-based approaches.

\textbf{Full corpus metaheuristic optimisation.} Extending the Firefly Algorithm search to the 9{,}253-sentence training set, possibly with a larger hyperparameter search space (e.g.\ maximum depth, minimum samples per leaf, feature sampling rate), could push macro-$F_1$ beyond the $0.9845$ mark already achieved on the sample.

\textbf{Active learning for minority classes.} Given the disproportionate impact of \textit{ideological} recall on macro-averaged metrics, an active-learning loop that prioritises annotation of uncertain minority-class candidates would be a cost-effective strategy for improving overall performance without annotating the entire unlabelled corpus.

\section{Conclusion}

This paper has presented a transparent and reproducible workflow for linking dependency-based syntactic annotation with sentence-level stylistic classification in Slovenian literary prose. Starting from Universal Dependencies parsing, we derived stylistic labels for 9{,}253 sentences across five narrative categories (\textit{descriptive}, \textit{introspective}, \textit{causal}, \textit{ideological}, \textit{neutral}) using an interpretable rule-based procedure, and then encoded each sentence as a linearised string of lemma--UPOS--dependency triples for machine-learning classification.

The experimental results establish a clear hierarchy among the evaluated approaches. Among the three feature-based classifiers, Random Forest consistently outperformed Logistic Regression and Linear SVC under all conditions --- the 3{,}300-sentence sample, the full 9{,}253-sentence corpus, and metaheuristic hyperparameter search --- reaching macro-$F_1 = 0.9764$ when trained on the full dataset and $0.9845$ after Firefly Algorithm optimisation with L\'{e}vy-flight perturbation. The cross-validation and $k$-fold sensitivity analyses confirm that these gains are stable and are not artefacts of any particular data partition.

A recurring challenge across all classifiers was the \textit{ideological} category, which accounts for fewer than $3.5\%$ of the corpus and exhibits partial syntactic overlap with \textit{descriptive} and \textit{causal} constructions. This minority-class difficulty underscores the importance of corpus balance and suggests that targeted data collection for underrepresented stylistic categories would yield the greatest return in future annotation efforts.

In contrast to the feature-based models, the DistilBERT architecture fine-tuned with LoRA adapters failed to match even the weakest baseline. We traced this failure to a fundamental input-format mismatch: the pipe-delimited syntactic triple strings are opaque to a model pre-trained on natural-language text, causing the subword tokeniser to fragment tokens in ways that destroy the symbolic structure on which the task depends. This finding carries a general methodological lesson: transfer learning is most beneficial when the pre-training distribution and the target input representation are closely aligned; when they diverge substantially, well-engineered symbolic features can outperform neural architectures by a considerable margin.

Several limitations should be kept in mind when interpreting these results. The corpus is restricted to a single literary domain, and it is unclear how well the learned TF-IDF vocabulary would generalise to journalistic, scientific, or conversational registers. Furthermore, the metaheuristic optimisation experiments were conducted on the 3{,}300-sentence sample rather than on the full dataset, so the optimised and full-dataset results are not directly comparable. Finally, the rule-based annotation provides a consistent and inspectable operationalisation of style, but it does not constitute a gold-standard human annotation; the classifiers learn to replicate a particular syntactic heuristic rather than human stylistic intuition.

Despite these limitations, the workflow demonstrates that even relatively coarse syntactic patterns --- restricted to lemma, POS tag, and dependency relation --- carry sufficient signal to discriminate between five functionally distinct narrative categories with high reliability. The method is intentionally lightweight: it requires only standard Python libraries, open-source parsing tools, and corpora in CoNLL-U format, making it accessible to researchers in digital humanities without specialised infrastructure. Because both the rule-based labelling logic and the TF-IDF feature space remain interpretable, the pipeline invites qualitative inspection alongside quantitative evaluation --- a property that is particularly valuable in literary and linguistic studies.

Three directions stand out for future work. First, replacing syntactic triples with raw sentence text as input to a Slovenian or South Slavic language model (e.g.\ SloBERTa) would align the pre-training distribution with the target domain and is expected to close the gap between neural and feature-based approaches. Second, extending the Firefly Algorithm search to the full 9{,}253-sentence training set, and widening the hyperparameter space to include tree depth and feature-sampling rate, is likely to push performance beyond the $0.9845$ macro-$F_1$ mark already achieved on the sample. Third, an active-learning loop that prioritises the annotation of uncertain minority-class candidates --- in particular \textit{ideological} --- would be a cost-effective strategy for improving macro-averaged performance without annotating the entire unlabelled corpus.

Taken together, the results suggest that dependency-based syntactic representation, combined with ensemble classification and metaheuristic optimisation, provides a robust and highly competitive baseline for stylistic sentence classification. We hope the workflow, as released with the accompanying code and data, will serve as a reusable foundation for further quantitative exploration of narrative style in Slovenian and other morphologically rich languages.

\bibliographystyle{apalike}
\bibliography{references}

\end{document}